\documentclass[runningheads]{llncs}

\usepackage{graphicx}
\usepackage[utf8]{inputenc}
\usepackage{url}
\usepackage{hyperref}
\usepackage{amssymb}
\usepackage{booktabs}
\hypersetup{
    colorlinks,
    citecolor=blue,
    filecolor=blue,
    linkcolor=blue,
    urlcolor=blue
}
\usepackage{cite}
\usepackage{amsmath,amssymb,amsfonts}
\usepackage{algorithm}
\usepackage{algpseudocode}
\usepackage{graphicx}
\usepackage{textcomp}
\usepackage{xcolor}
\usepackage{url}
\usepackage{bbold}
\usepackage{bm}
\usepackage{bbm}
\usepackage[euler]{textgreek}
\usepackage{listings}

\DeclareMathOperator*{\bx}{{\mathbf{x}}}

\DeclareMathOperator*{\bs}{{\mathbf s}}

\DeclareMathOperator*{\btheta}{{\bm{\theta}}}

\graphicspath{{./fig/}}

\begin{document}

\title{Error-guided likelihood-free MCMC}

\titlerunning{Error-guided likelihood-free MCMC}

\author{Volodimir Begy\inst{1,2} \and
Erich Schikuta\inst{1}}

\authorrunning{V. Begy et al.}

\institute{University of Vienna, Faculty of Computer Science,
Research Group Workflow Systems and Technology,
W{\"a}hringer Stra{\ss}e 29, 1090, Vienna, Austria\\
\and
CERN, 1211 Geneva 23, Switzerland\\
\email{volodimir.begy@cern.ch}}

\maketitle

\begin{abstract}

 This work presents a novel posterior inference method for models with intractable evidence and likelihood functions.
 Error-guided likelihood-free MCMC, or EG-LF-MCMC in short, has been developed for scientific applications, where
 a researcher is interested in obtaining approximate posterior densities over model parameters, while avoiding
 the need for expensive training of component
 estimators on full observational data or the tedious design of expressive summary statistics, as in related approaches.
 Our technique is based on two phases. In the first phase, we
 draw samples from the prior, simulate respective observations and record their errors $\epsilon$ in relation
 to the true observation. We train a classifier to distinguish between corresponding and non-corresponding
 ($\epsilon, \btheta$)-tuples. In the second stage the said classifier is conditioned on the smallest recorded
 $\epsilon$ value from the training set and employed for the calculation of
 transition probabilities in a Markov Chain Monte Carlo sampling procedure. By conditioning the MCMC on specific
 $\epsilon$ values, our method may also be used in an amortized fashion to infer posterior
 densities for observations, which are located a given distance away from the observed data.
 We evaluate the proposed method on benchmark problems with semantically and structurally different data and
 compare its performance against the state of the art
 approximate Bayesian computation (ABC).

\end{abstract}

\section{Motivation}

In natural sciences the studied phenomena are typically formalized by mathematical models.
These models may come in various flavors, ranging from systems of differential equations to
complex stochastic computer simulations.
With the advancement of computing technologies it has become increasingly common to specify mechanistic models
using complex computer programs. This gives rise to the negative side effect that the likelihood functions and the model evidence
are typically intractable, meaning, they have no closed-form expressions due to the model complexity.
Researchers often wish to evaluate theories or refine the developed statistical models.
An applicable tool for this is posterior inference, whereby a practitioner seeks a conditional
probability density function over model parameters given observed data.
In modern scientific applications, posterior inference is a non-trivial task. As an example,
consider a computer science researcher, who has experimentally gathered some metrics describing the performance
of a data grid, which is a large-scale distributed data-intensive computing system. This production
system is abstracted by a simulator with numerous, potentially latent parameters. The researcher would like
to relate the observed data $\bx$ to concrete values of the simulator parameters $\btheta$, by obtaining a posterior
$p(\btheta \vert \bx)$.
Due to the intractability of the likelihood function $p(\bx \vert \btheta)$ and model evidence $p(\bx)$, this task cannot be performed analytically
by the means of the Bayes' theorem:

\begin{equation}
p(\btheta \vert \bx) = \frac{p(\bx \vert \btheta) p(\btheta)}{p(\bx)}.
\end{equation}

This issue has motivated the field of likelihood-free or simulation-based
inference \cite{Cranmer201912789}, in which researchers develop new inference methods, which do not rely on the
direct evaluation of likelihood functions and model evidence. Typically, such novel inference techniques
are developed using tools from soft computing and statistics.

We propose a method for the construction of the error-conditioned posterior $p(\btheta \vert \epsilon)$ instead of
the traditional posterior $p(\btheta \vert \bx)$. Error-guided likelihood-free MCMC can be used for 2 different
scientific purposes by employing a single trained classifier in an amortized fashion.
Firstly, it allows to obtain the posterior $p(\btheta \vert \epsilon_{min})$, where $\epsilon_{min}$ is the smallest recorded
error from the training set in relation to the true observation $\bx_o$. The parameters $\btheta$ drawn from
$p(\btheta \vert \epsilon_{min})$ aim at
reproducing the observed data $\bx_o$. Secondly, the practitioner may be interested in deriving posterior densities over model
parameters, which allow to generate observations a given distance away from the observed data. This is achievable using our
approach by conditioning the MCMC procedure on any $\epsilon$ value, which is covered by the range of the training set.

\section{Related Work}

Arguably the most popular method for likelihood-free posterior sampling is approximate Bayesian computation (ABC) \cite{marin2012approximate, Beaumont2025}.
Despite being conceptually simple, it is relatively effective. In rejection ABC a certain $\btheta$ is drawn from the prior $p(\btheta)$ and
a respective observation $\bx \sim\ p(\bx|\btheta)$ gets simulated. Then, a selected distance function $d(\bx, \bx_o)$
calculates the distance between the observed and simulated data. The observations may also be compressed into lower
dimensions by summary statistics. The construction of expressive summary statistics is a non-trivial
task \cite{doi:10.1111/j.1467-9868.2011.01010.x, Aeschbacher1027, OnOptimalSelectionofSummaryStatisticsforApproximateBayesianComputation}.
In case the distance is below a selected
threshold $\epsilon$, the generating parameter $\btheta$ is accepted. This procedure is repeated, until the collected
sample is large enough to approximate the posterior density. Typically, ABC is relatively inefficient in high-dimensional
$\btheta$-spaces, since most simulated observations get rejected. We note that in the ABC literature $\epsilon$ stands
for the acceptance threshold, while in our formalism $\epsilon$ denotes concrete distances between the candidate
and true observations.

In the recent years, plenty of more advanced likelihood-free inference methods have emerged. Our work is inspired by
Likelihood-free MCMC with Amortized Approximate Ratio Estimators, or AALR-MCMC \cite{hermans2019likelihood}.
AALR-MCMC trains an artificial neural network classifier to learn a mapping between the $\btheta$ and $\bx$ spaces
by distinguish corresponding ($\bx, \btheta$)-tuples from non-corresponding ones. In case of high dimensionality
of the $\bx$ space, the observed data can be compressed by a summary statistic. It is demonstrated that an optimally
trained classifier outputs $\frac{p(\bx, \btheta)}{p(\bx, \btheta)+p(\bx)p(\btheta)}$, allowing us to calculate the
likelihood-to-evidence ratio for a given input pair $(\bx, \btheta)$. In order to obtain a posterior,
an MCMC sampling scheme is used. While constructing a Markov Chain, the probability of the transition from
the current state $\btheta_t$ to a proposed state $\btheta'$ is calculated using the estimated likelihood ratio
$\hat{r}(\bx|\btheta_t, \btheta')$, which is extracted from the converged classifier. A major strength of this technique
is its ability to perform amortized inference for different observations $\bx$ using a single trained classifier.

A family of approaches constructs the approximate prior using sequential procedures \cite{2018arXiv181108723D}.
In SNPE-A \cite{papamakarios2016fast} the authors refine a proposal prior $\tilde{p}(\bx)$ by iteratively re-training
a parameterized conditional density estimator $q_{\phi}(\btheta|\bx)$, which is realized using single-component
Mixture Density Networks \cite{bishop1994mixture}. Once $\tilde{p}(\bx)$ converges, the final approximate posterior
is derived after training an MDN with multiple components. SNPE-B \cite{NIPS2017_6728} extends upon this work by
training the MDN using an importance-weighted loss function. This allows $q_{\phi}(\btheta|\bx)$ to output
posterior densities without additional analytical transformations. However, the authors in \cite{pmlr-v97-greenberg19a} argue,
that such importance-weighting gives rise to a high variance of the density estimator parameters.
Furthermore, the loss function includes a term, which calculates the Kullback-Leibler divergence between the
distributions over MDN parameters from the
previous and the current optimization rounds. Thus, the optimization procedure primarily focuses on refining
highly uncertain parameters. The authors improve the simulation efficiency by discarding proposed $\btheta$,
which are deemed as nonsensical by a binary classifier. SNPE-C/APT \cite{pmlr-v97-greenberg19a} optimizes
a conditional estimator, which directly targets the posterior, by minimizing the estimated negative
log density of a proposal posterior over samples drawn from a proposal prior. SNL \cite{pmlr-v89-papamakarios19a}
opts for learning an estimator of the likelihood function $q_{\phi}(\bx|\btheta)$ and obtains the approximate posterior
at round $r$ by $\hat{p_r} \propto q_{\phi}(\bx_o|\btheta)p(\btheta)$. Hermans et al. \cite{hermans2019likelihood}
demonstrate that AALR-MCMC can also be executed as a sequential procedure by using the approximated posterior
from the round $r$ as the prior in the round $r+1$. Such implementation makes sense when the capacity of
the classifier is not sufficient for it to converge in a single round.

Durkan et al. \cite{durkan2020contrastive} have shown that AALR-MCMC and SNPE-C are instances of contrastive learning
and presented a generalized algorithm for Sequential Contrastive Likelihood-free Inference.

\section{The Method}

EG-LF-MCMC consists of two phases.
Firstly, we
draw $\btheta$ samples from the prior $p(\btheta)$,
generate observations $\bx \sim\ p(\bx|\btheta)$ from the implicit model and record their errors $\epsilon$ in relation
to the true observation $\bx_o$ by applying a distance function $d(\bx,\bx_o)$.
We train a classifier $\bs(\epsilon,\btheta)$ to distinguish between corresponding and non-corresponding
($\epsilon, \btheta$)-tuples. In the second stage the output of the classifier is transformed into estimated
ratios $\hat{r}(\epsilon|\btheta',\btheta_t)$
in order to evaluate the
transition probabilities for subsequent state samples $\btheta'$ and $\btheta_t$ in an MCMC sampling scheme.
The method may be run iteratively, by setting the posterior $p(\btheta|\epsilon)$ from the round $r$ as the prior
at the round $r+1$.

\begin{algorithm}[H]
  \caption{Training of $\bs(\epsilon,\btheta)$}
  \label{algo:train}
  \begin{tabular}{ l l }
    {\it Inputs:} & Generative model $p(\bx \vert \btheta)$ \\
                  & Prior $p(\btheta)$ \\
                  & True observation $\bx_o$ \\
                  & Distance function $d(\bx,\bx_o)$ \\
    {\it Outputs:} & Classifier $\bs_\phi(\epsilon,\btheta)$ \\
    {\it Hyperparameters:} & Batch-size $M$ \\
  \end{tabular}
  \\
  \begin{algorithmic}[1]
    \While{$\bs_\phi(\epsilon,\btheta)$ \bf not converged}
    \State Sample $\btheta \gets \{\btheta_m \sim p(\btheta)\}_{m=1}^{M}$
    \State Sample ${\btheta}^{'} \gets \{{\btheta}_m^{'} \sim p(\btheta)\}_{m=1}^{M}$
    \State Simulate $\bx \gets \{{\bx}_m \sim p(\bx\vert\btheta_m)\}_{m=1}^{M}$
    \State Simulate ${\bx}{'} \gets \{{\bx}'_m \sim p(\bx\vert{\btheta}_m^{'})\}_{m=1}^{M}$
    \State $\epsilon \gets \{d({\bx}_m, \bx_o)\}_{m=1}^{M}$
    \State $\epsilon^{'} \gets \{d({\bx}'_m, \bx_o)\}_{m=1}^{M}$
    \State $\mathcal{L}_a \gets \textsc{BCE}(\bs_\phi(\epsilon,\btheta),~\mathbb{1}) + \textsc{BCE}(\bs_\phi({{\epsilon}{'}},{\btheta}),~\mathbb{0})$
    \State $\mathcal{L}_b \gets \textsc{BCE}(\bs_\phi({{\epsilon}{'}},{\btheta}{'}),~\mathbb{1}) + \textsc{BCE}(\bs_\phi({\epsilon},{\btheta}{'}),~\mathbb{0})$
    \State $\mathcal{L} \gets \mathcal{L}_a + \mathcal{L}_b$
    \State $\phi \gets \textsc{optimizer}(\phi,~\nabla_\phi\mathcal{L})$
    \EndWhile
    \State \Return{$\bs_\phi(\epsilon,\btheta)$}
  \end{algorithmic}
\end{algorithm}

Our method builds upon AALR-MCMC and modifies it in the following ways: (1) the classifier is trained on
errors $\epsilon$ in relation to $\bx_o$ instead of original or compressed observational data $\bx$; (2) the MCMC procedure
is conditioned on $\epsilon$ instead of $\bx$. When inferring a posterior $p(\btheta|\epsilon_{min})$,
which aims at reproducing $\bx_o$, we pass
the smallest simulated $\epsilon$ value to the MCMC. Alternatively, if one wishes to infer a posterior for observations,
which are a given distance away from $\bx_o$, it is possible to specify any $\epsilon$ value, which is covered by the range
of the training set.

The function $d(\bx,\bx_o) = \epsilon$ can be flexibly specified taking into account the semantics of the
studied problem. We have found the $L_{1}$ distance to work well on a wide range of problems:

\begin{equation}
d(\bx,{\bx}_o) = \sum_{i=1}^{N}{\lvert \bx - {\bx}_o \rvert}_i
\end{equation}

Training the classifier on $\epsilon$ brings several advantages. In problems with high-dimensional
observations, the size of the training set becomes drastically reduced, resulting into memory and computational
efficiency. It becomes easier to specify a sufficient classifier architecture.
These advantages can also be obtained by compressing the data using a summary statistic. However, we argue that
it is easier to specify a semantically viable distance function than an expressive summary statistic for
non-trivial high-dimensional data. Lastly, the novel amortization over $\epsilon$ may come handy in specific
scientific use-cases.

The training procedure for the classifier $\bs(\epsilon,\btheta)$ is adapted from AALR-MCMC and is presented
in the Algorithm~\ref{algo:train},
where BCE stands for the binary cross entropy loss function. In \cite{hermans2019likelihood} the authors have shown that
the classifier trained by AALR-MCMC to optimality outputs $\frac{p(\bx, \btheta)}{p(\bx, \btheta)+p(\bx)p(\btheta)}$ for a given
$(\bx, \btheta)$-pair and that this output can be transformed into the likelihood-to-evidence ratio:

\begin{equation}
\frac{p(\bx|\btheta)}{p(\bx)}=r(\bx|\btheta)
\end{equation}

Since our procedure replaces $\bx$ by $\epsilon$, a classifier optimally trained by EG-LF-MCMC outputs
$\frac{p(\epsilon, \btheta)}{p(\epsilon, \btheta)+p(\epsilon)p(\btheta)}$, which is transformed into the ratio:

\begin{equation}
\frac{p(\epsilon|\btheta)}{p(\epsilon)}=r(\epsilon|\btheta)
\end{equation}

In order to perform the MCMC sampling, we adapt the
likelihood-free Metropolis-Hastings sampler \cite{metropolis1953equation, 10.1093/biomet/57.1.97} from AALR-MCMC
for the use of our classifier $\bs(\epsilon,\btheta)$, as summarized
by the Algorithm~\ref{algo:MCMC}. We note that in the common case of a uniform prior the terms $\log p(\btheta{'})$ and
$\log p({\btheta}_t)$ from the 5th line can be removed. The term
$\frac{q({\btheta}_t \vert {\btheta}')}{q({\btheta}' \vert {\btheta}_t)}$
from the 6th line can be removed in case of a symmetric transition distribution.
These modifications will speed up the computation.

\begin{algorithm}[H]
  \caption{Error-guided likelihood-free Metropolis-Hastings}
  \label{algo:MCMC}
  \begin{tabular}{ l l }
    {\it Inputs:} & Initial parameter $\btheta_0$ \\
                  & Prior $p(\btheta)$ \\
                  & Transition distribution $q(\btheta)$ \\
                  & Classifier $\bs(\epsilon, \btheta)$ \\
                  & Error $\epsilon$ \\
    {\it Outputs:} & Markov chain $\btheta_{0:n}$ \\
    {\it Hyperparameters:} & Markov chain size $n$ \\
  \end{tabular}
  \\
  \begin{algorithmic}[1]
    \State{$t \gets 0$}
    \State{$\btheta_t \gets \btheta_0$}
    \For{$t < n$}
    \State $\btheta' \sim q(\btheta \vert \btheta_t)$
    \State $\displaystyle\lambda \gets (\log\hat{r}(\epsilon\vert\btheta{'}) + \log p(\btheta{'})) - (\log\hat{r}(\epsilon\vert{\btheta}_t) + \log p({\btheta}_t))$
    \State $\displaystyle\rho \gets \min(\exp(\lambda) \frac{q({\btheta}_t \vert {\btheta}')}{q({\btheta}' \vert {\btheta}_t)},~1)$
    \State $\btheta_{t + 1} \gets \begin{cases}
    \btheta' & \text{with probability } \rho \\
    \btheta_t & \text{with probability } 1 - \rho
    \end{cases}$
    \State $t \gets t + 1$
    \EndFor
    \State \Return{$\btheta_{0:n}$}
  \end{algorithmic}
\end{algorithm}

\section{Experiments}

We compare EG-LF-MCMC to rejection ABC, since both approaches are similar in that they operate on calculating a distance
between a candidate observation $\bx$ and the true observation $\bx_o$. In particular, we evaluate the ability of
$p(\btheta|\epsilon_{min})$ from EG-LF-MCMC and $p(\btheta|\bx)$ from ABC to recover the true generating parameter $\btheta^*$,
which gives rise to the true observation $\bx_o$.

\begin{table}[ht]
\begin{center}
\begin{tabular}{llll}\toprule
  &&\textbf{Circle} & \textbf{Linear}\\\midrule
  \textbf{EG-LF-MCMC} & \textit{n simulations} & 12,033,424 & 12,652,661\\
  & \textit{n training epochs} & 76 & 37\\
  \textbf{ABC} & \textit{n simulations} & 134,582,382 & 298,734,182\\
  & $\epsilon$ & 100 & 200
  \\\bottomrule
\end{tabular}
\end{center}
\caption{Further implementation details}
\label{tab:metrics}
\end{table}

\begin{figure}
\centering
\begin{minipage}{.5\textwidth}
  \centering
  \includegraphics[width=\textwidth]{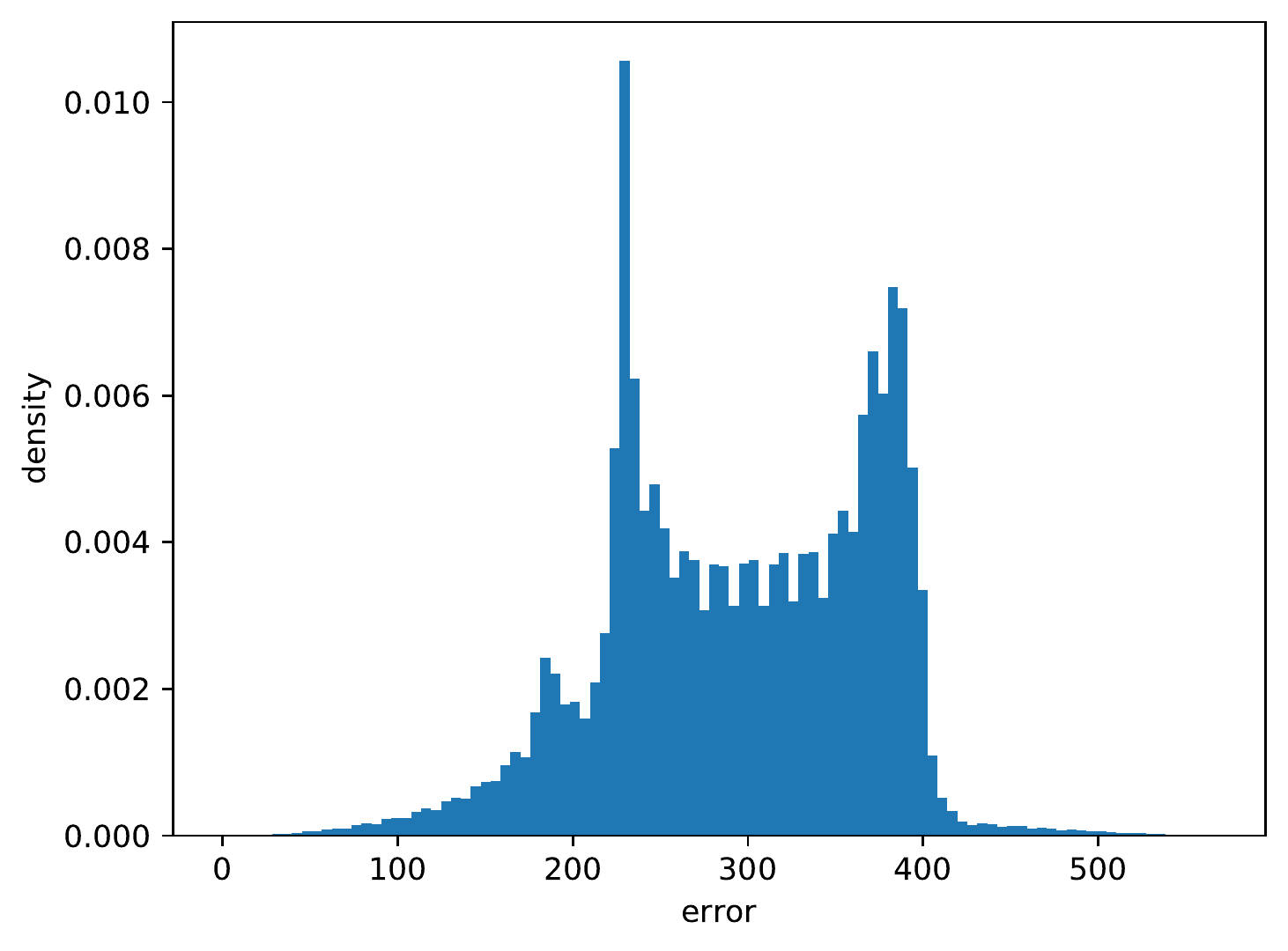}
\end{minipage}%
\begin{minipage}{.5\textwidth}
  \centering
  \includegraphics[width=\textwidth]{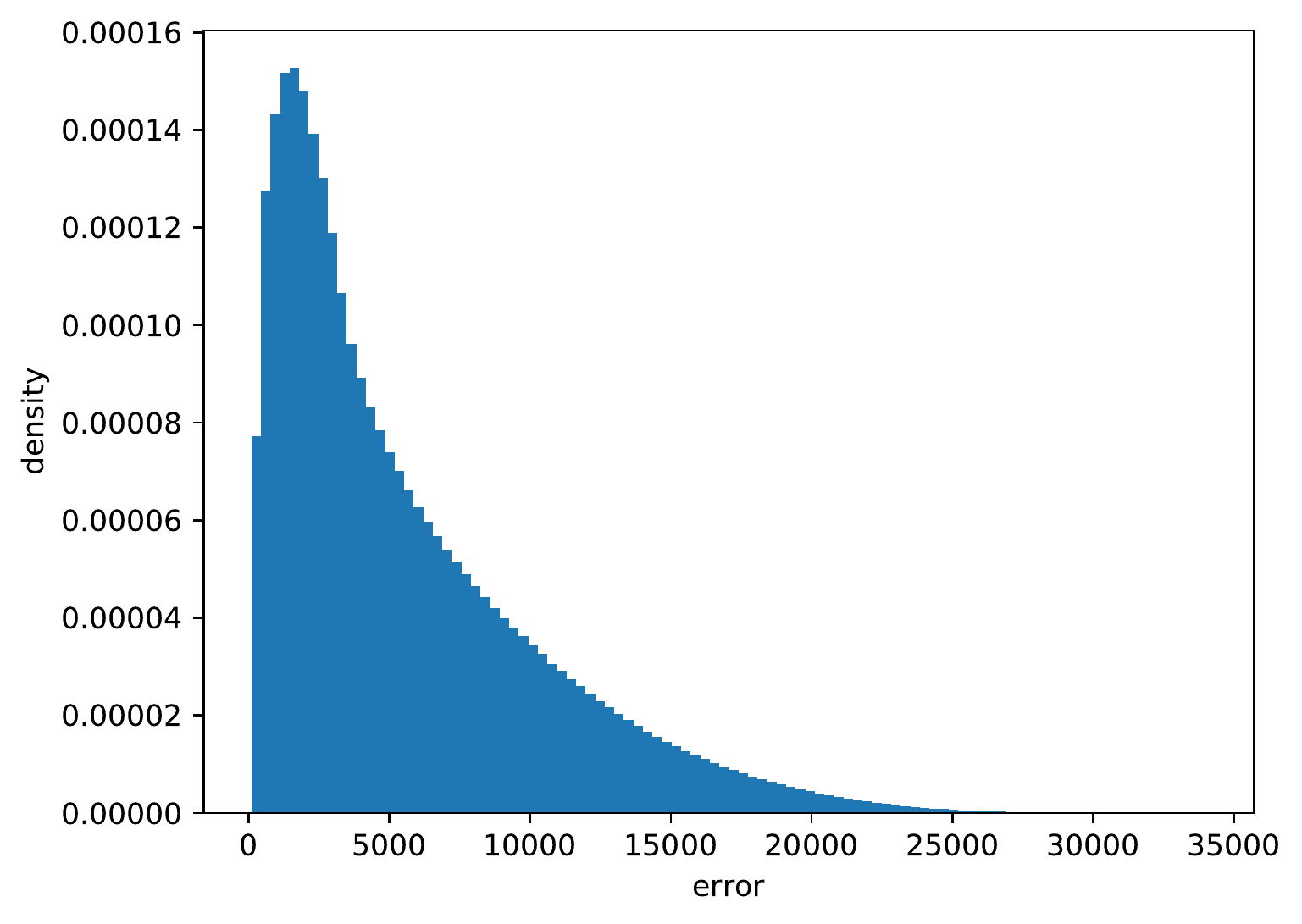}
\end{minipage}
\caption{Distribution of $\epsilon$ for the circle (left) and the linear (right) benchmark problems}
\label{fig:error}
\end{figure}

On each task, both methods employ the $L_1$ distance to
calculate the discrepancy between the candidate and the true observations.
The EG-LF-MCMC classifiers are realized using a
multilayer perceptron with 4 hidden layers, each having 128 hidden units and SELU \cite{NIPS2017_6698} non-linearities.
The nets are trained using the Adam optimizer \cite{2014arXiv1412.6980K} with a learning rate of 0.0001 and a batch size of 64.
We project the training data onto the [0, 1] interval, as the experiments have indicated that this stabilizes the training.
Furthermore, after such projection, the MCMC procedure can always be conditioned on ${\epsilon}_{min}=0$, without the extra
effort of scanning the training set for the actual value of the smallest recorded $\epsilon$.
We use the PyTorch framework \cite{NEURIPS2019_9015} for the implementation and the training of the neural nets.
Both methods draw 1 million samples in order to approximate
the posterior. EG-LF-MCMC additionally performs 10,000 burn-in steps.
Other variable implementation metrics for both methods are reported in the Table \ref{tab:metrics}.
Since both approaches share a single prior, we pick the value for the acceptance threshold in
ABC based on the analysis of the distributions of $\epsilon$ from the training set of EG-LF-MCMC, which
are depicted in the Fig. \ref{fig:error}.

\begin{figure}
\centering
\begin{minipage}{.42\textwidth}
  \centering
  \includegraphics[width=\textwidth]{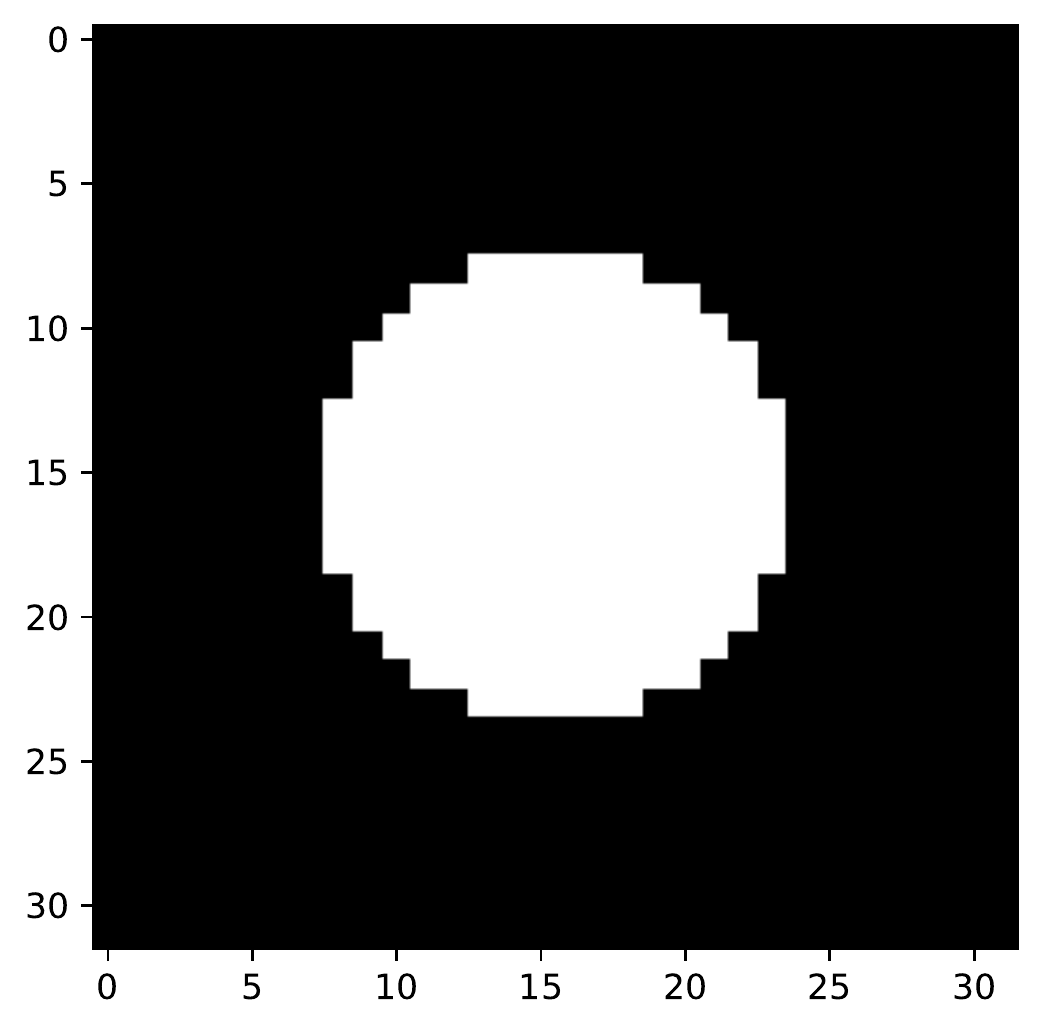}
\end{minipage}%
\begin{minipage}{.58\textwidth}
  \centering
  \includegraphics[width=\textwidth]{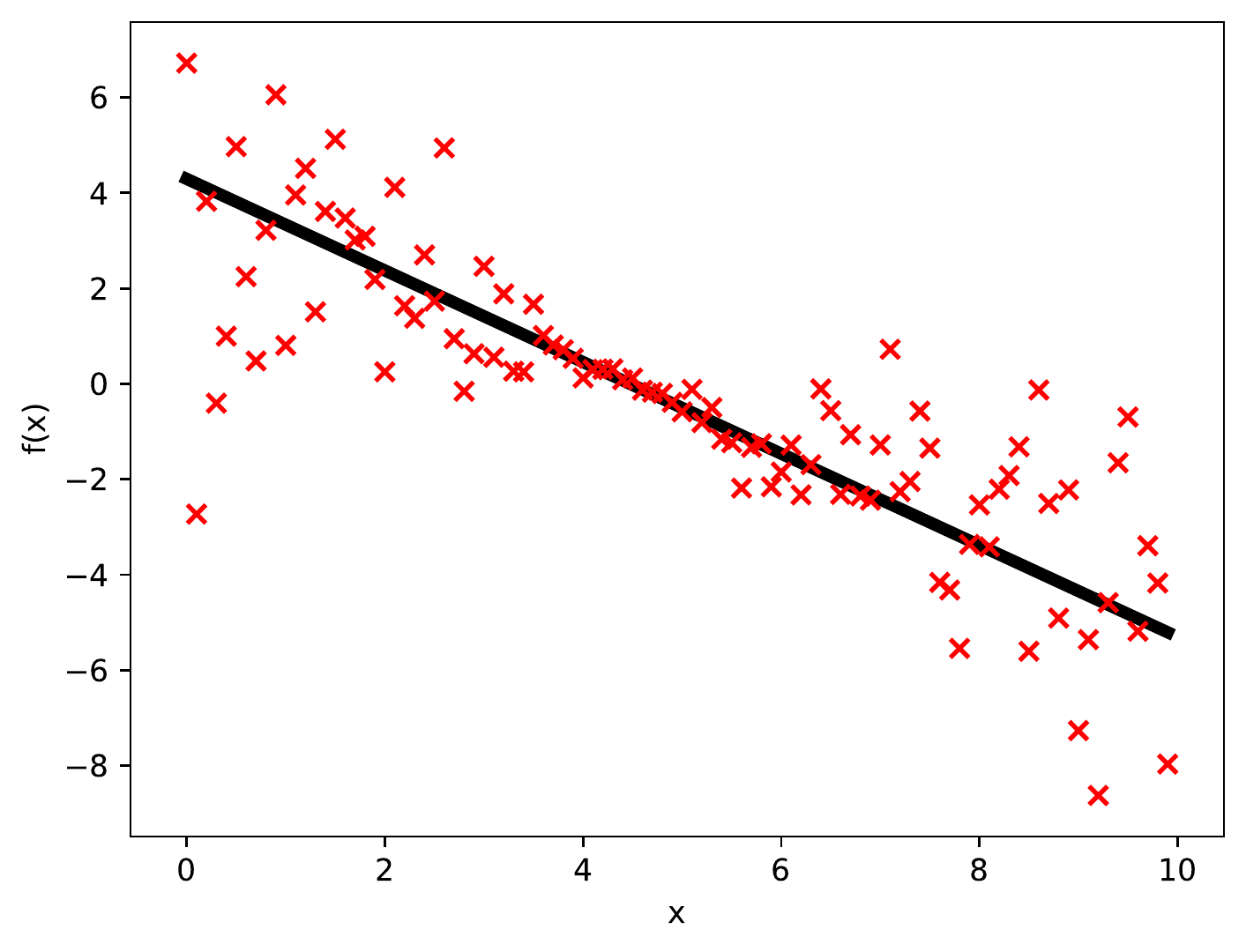}
\end{minipage}
\caption{Example $\bx_o$ for the circle (left) and the linear (right) benchmark problems}
\label{fig:xo}
\end{figure}

\subsubsection{Circle image simulator}

\begin{figure}
\centering
\begin{minipage}{.5\textwidth}
  \centering
  \includegraphics[width=\textwidth]{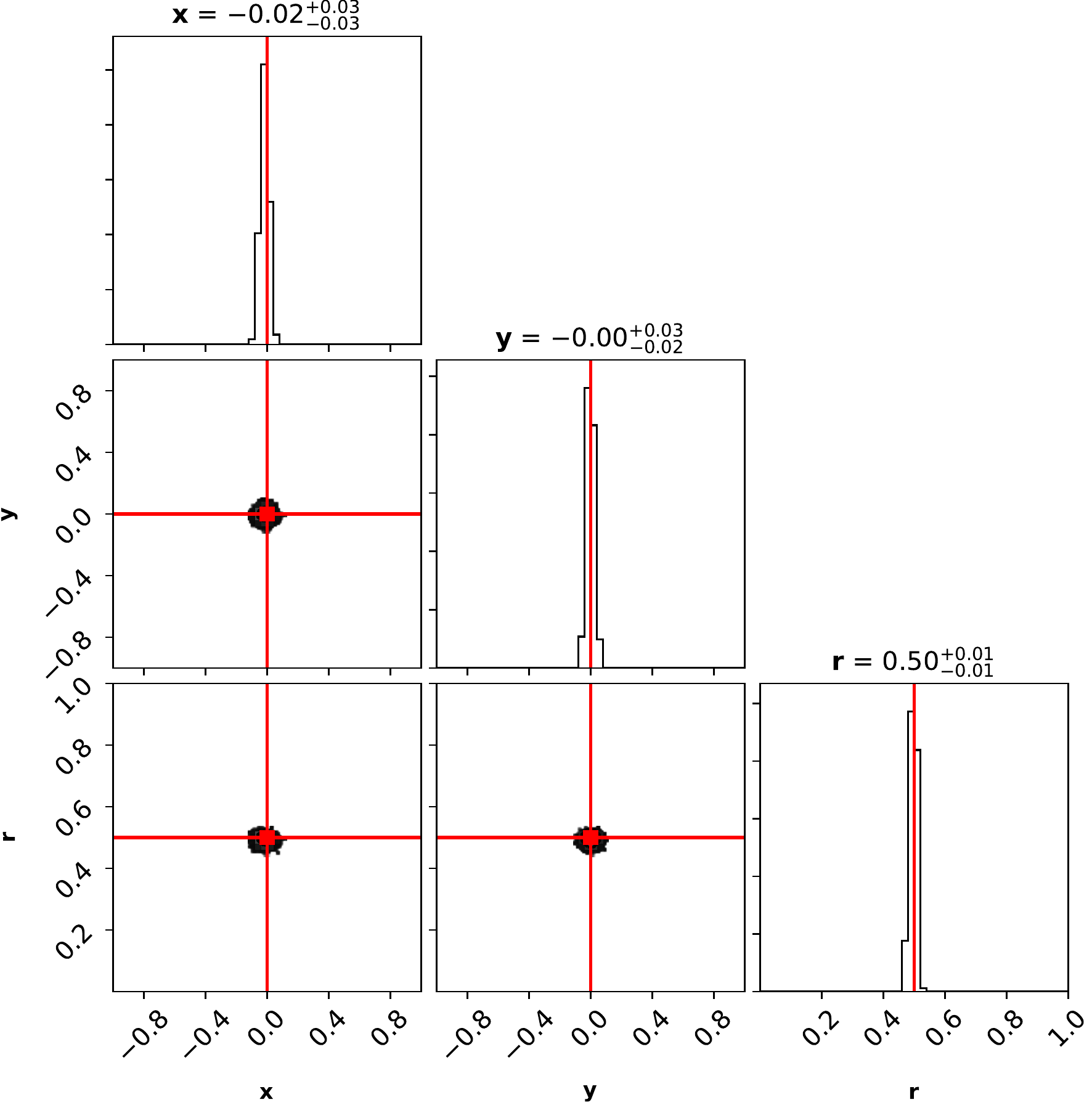}
\end{minipage}%
\begin{minipage}{.5\textwidth}
  \centering
  \includegraphics[width=\textwidth]{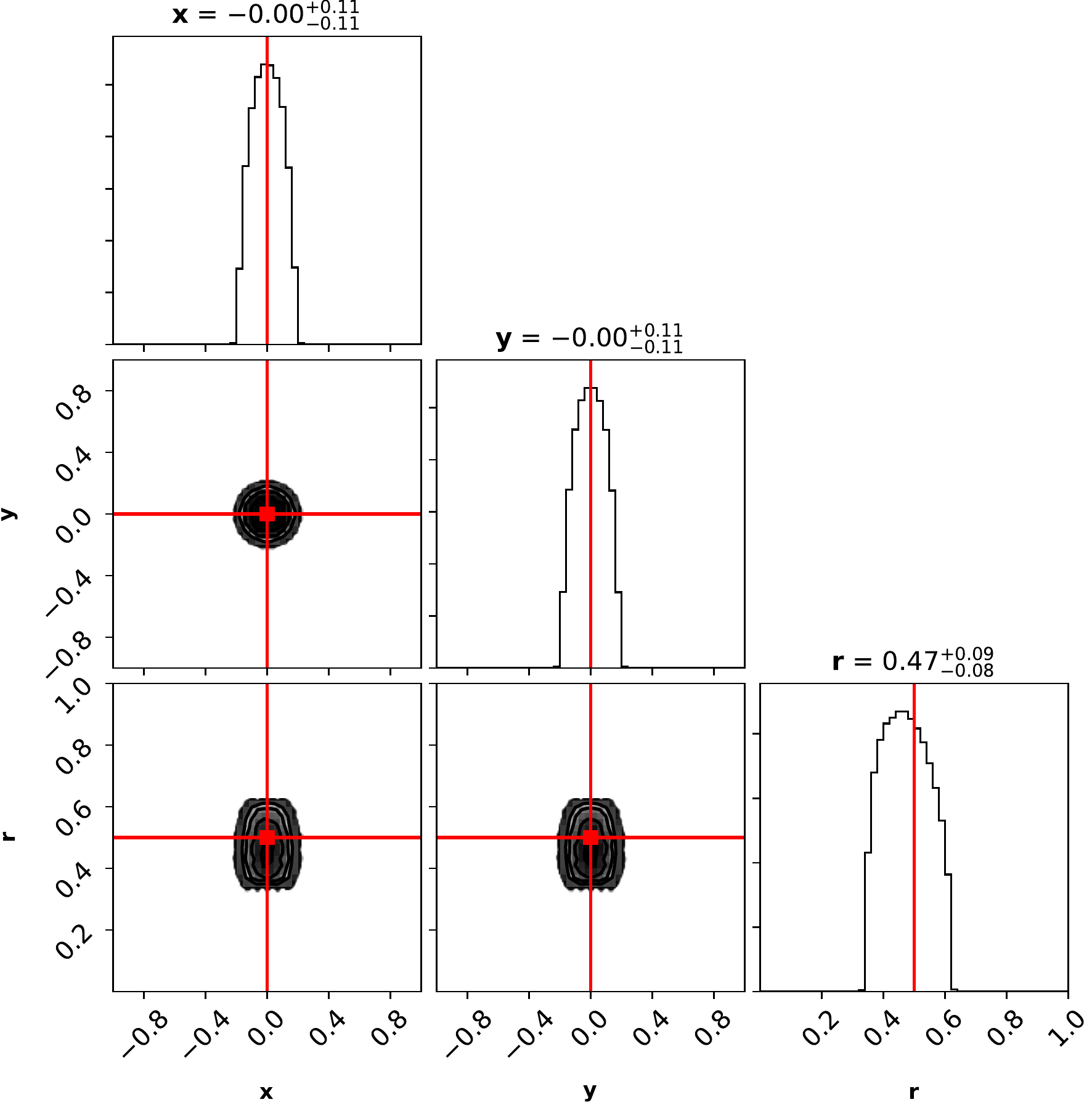}
\end{minipage}
\caption{Posteriors for the circle problem. The true generating parameter $\btheta^*$
is depicted in red. \textit{Left:} $p(\btheta|\epsilon)$ from EG-LF-MCMC.
\textit{Right:} $p(\btheta|\bx)$ from ABC}
\label{fig:posterior_circle}
\end{figure}

The first benchmark problem is based on a deterministic simulator of 2-dimensional circle images, while both
$x$ and $y$ coordinates are constrained by the interval [-1, 1] \cite{hypothesis}.
The images are black and white, having a size of 32x32 pixels. The simulator is configured by 3 parameters $\btheta=(x,y,r)$,
where $x$ and $y$ are the coordinates of the circle's center, and $r$ is its radius. The true observation
$\bx_o$ is generated from $\btheta^*=(x=0,y=0,r=0.5)$ and is depicted on the left side of the Fig. \ref{fig:xo}.
We assume a uniform prior with the range of [-1, 1] for $x$ and $y$ and the range of [0, 1] for $r$.
Certain simulated observations from the training set of EG-LF-MCMC are able to achieve
$\epsilon=0$ and exactly reproduce ${\bx}_o$.
The obtained posteriors are presented in the Fig. \ref{fig:posterior_circle}.

\subsubsection{Stochastic linear model} 
\begin{figure}
\centering
\begin{minipage}{.5\textwidth}
  \centering
  \includegraphics[width=\textwidth]{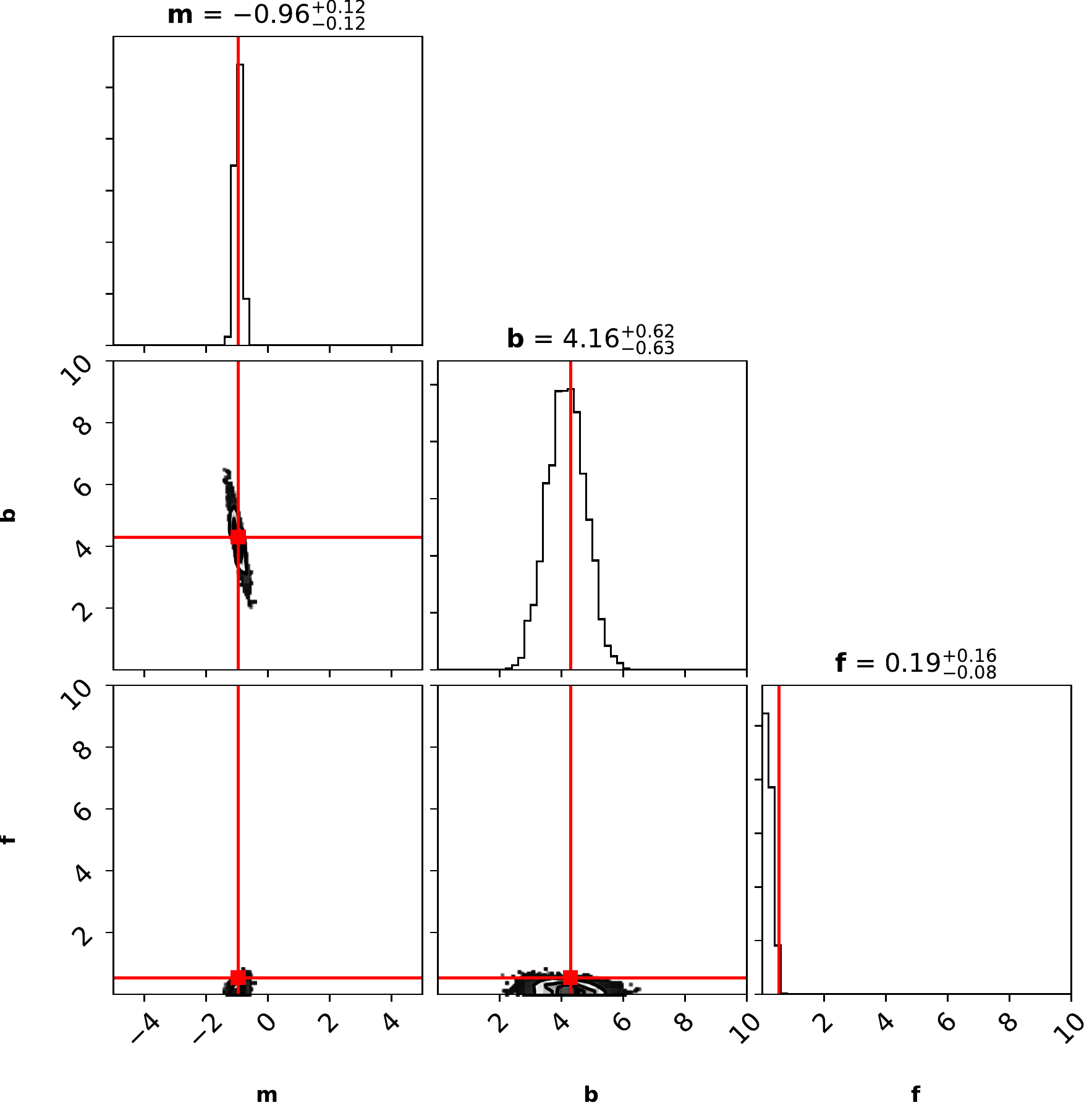}
\end{minipage}%
\begin{minipage}{.5\textwidth}
  \centering
  \includegraphics[width=\textwidth]{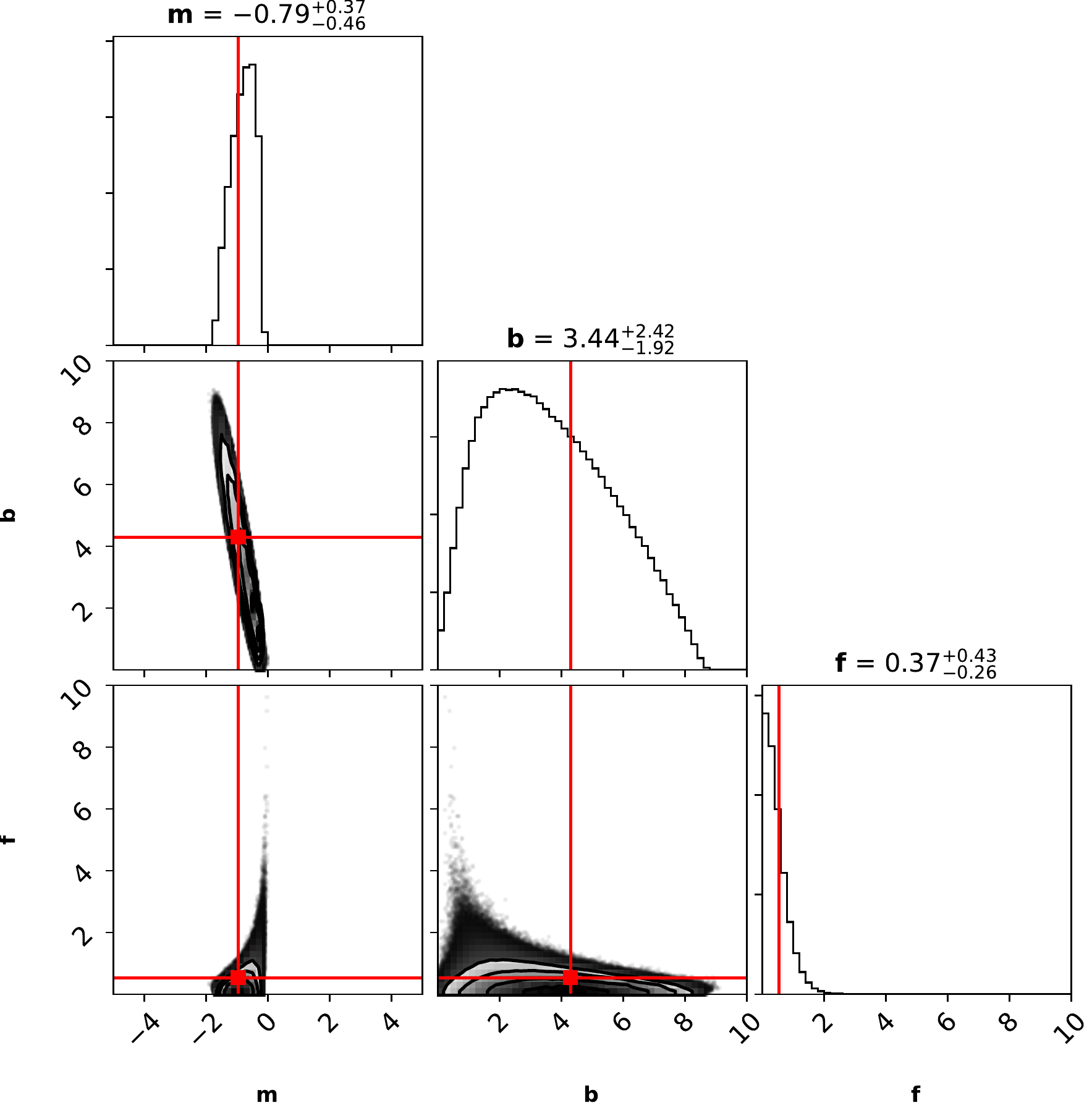}
\end{minipage}
\caption{Posteriors for the linear problem. The true generating parameter $\btheta^*$
is depicted in red. \textit{Left:} $p(\btheta|\epsilon)$ from EG-LF-MCMC.
\textit{Right:} $p(\btheta|\bx)$ from ABC}
\label{fig:posterior_linear}
\end{figure}

Secondly, we compare the two approaches on the benchmark problem
of a stochastic linear model, adapted from \cite{emcee, hermans2019likelihood}.
The model is specified as:

\begin{equation}
y = mx + b + {\epsilon}_1 \lvert f(mx + b) \rvert + 0.05{\epsilon}_2
\end{equation}

\noindent where ${\epsilon}_1$ and ${\epsilon}_2$ are drawn from the standard normal $\mathcal{N}(0,\,1)$.
In this model $mx + b$ calculates a deterministic value for $y$ based on a standard linear function.
The expression ${\epsilon}_1 \lvert f(mx + b) \rvert$ adds a stochastic error to it based on the magnitude of
$mx + b$. The term $0.05{\epsilon}_2$ adds a further error.
We generate a 100-dimensional observation $\bx_o$ by applying $\btheta^*=(m=-0.9594, b=4.294, f=0.534)$
and measuring the values $({y}_1, {y}_2, ..., {y}_{100})$ at 100 equidistant $x$ points. An example $\bx_o$ for this model
is depicted on the right side of the Fig. \ref{fig:xo}.
We assume a uniform prior with the range of [-5, 5] for $m$ and the range of [0, 10] for $b$ and $f$.
The smallest $\epsilon$ recorded by the training set of EG-LF-MCMC is 94.7.
The obtained posteriors are presented in the Fig. \ref{fig:posterior_linear}.

\subsubsection{Discussion} Despite requiring drastically less calls to the implicit generative model, as
can be seen from Table \ref{tab:metrics}, the posteriors obtained by EG-LF-MCMC are significantly more
certain about the true generating parameter $\btheta^*$ on both benchmark problems.
Concerning the linear problem, both approaches recover $f^*$, but tend to
underestimate it. We explain this behavior as follows. The term $f$ controls the magnitude of the stochastic
error in the observations. Since ${\epsilon}_1$ is drawn from a standard normal, the error
${\epsilon}_1 \lvert f(mx + b) \rvert$ can shift an observation's value ${y}_x$, where $1 \leq x \leq 100$,
in both positive and negative directions.
Thus, even if the true value $f^*$ is inferred, across the true observation ${\bx}_o$ and
a candidate observation $\bx$ the values ${y}_x$ may be stochastically shifted in opposite directions,
leading to a greater error. Since both EG-LF-MCMC and ABC try to infer the posteriors by minimizing the error,
it is safer to assume a smaller value for $f$. This way $\bx$ will stay closer to ${\bx}_o$ on average, across different
simulation runs.

\subsection{Amortization over $\epsilon$}
\begin{figure}
\centering
\includegraphics[width=.5\textwidth]{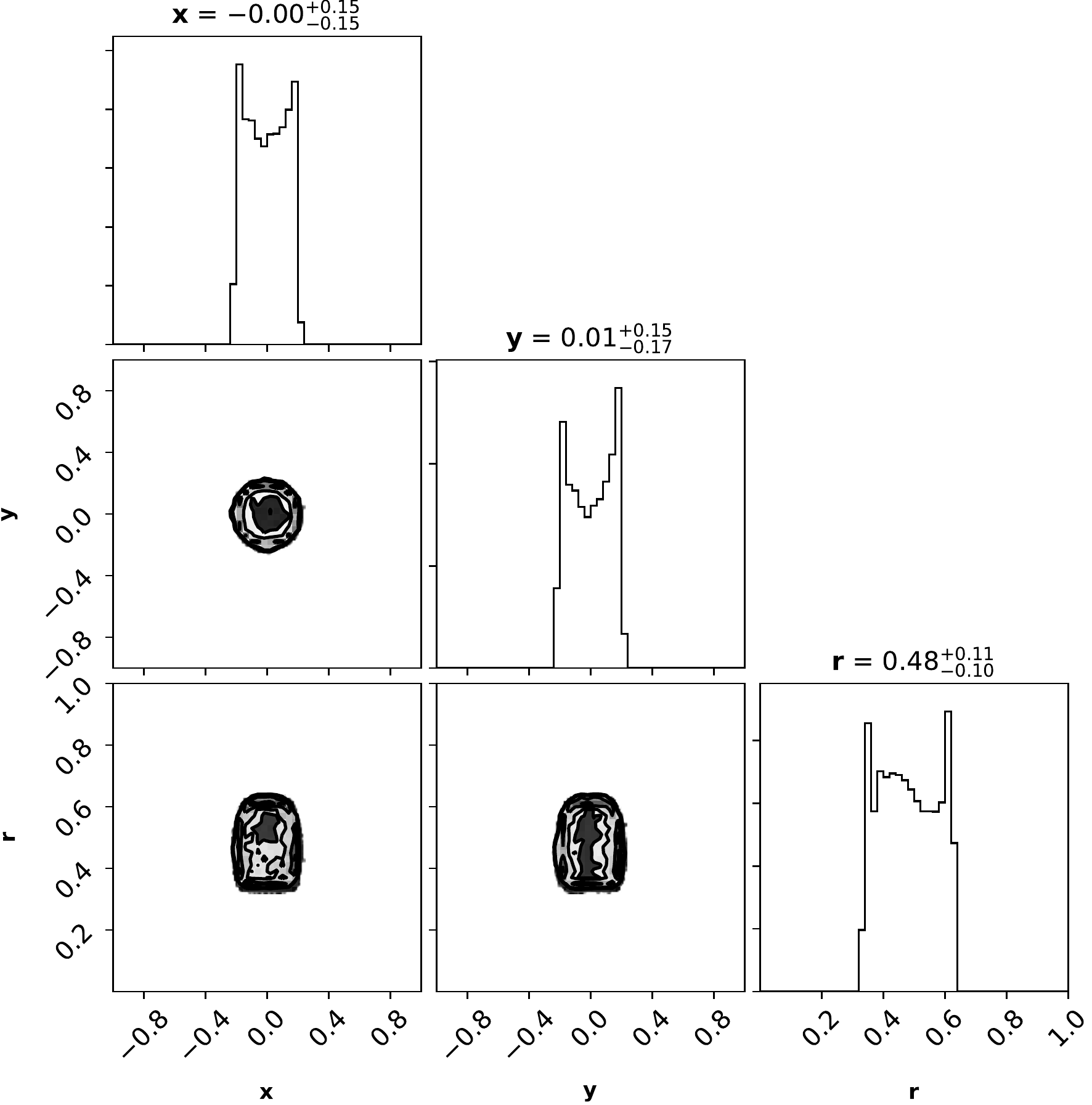}
\caption{Circle problem: approximate posterior density $p(\btheta|\epsilon = 100)$}
\label{fig:eps_posterior}
\end{figure}

We demonstrate the ability of EG-LF-MCMC to amortize the inference over $\epsilon$, which is
a completely new feature compared
to AALR-MCMC. For this purpose, we turn again to the circle problem.
We use the same trained classifier (amortization) as in the previously
presented inference for $\btheta^*$ in order to infer approximate
posterior densities for observations, which are a given distance $\epsilon$ away from $\bx_o$.
We denote such observations by $\bx_{\epsilon}$. Concretely, we would like $\bx_{\epsilon}$ to differ from
$\bx_o$ by 100 pixels. For this purpose, we condition the MCMC on $\epsilon = 100$.
The sampled multivariate chain is depicted in Fig. \ref{fig:eps_posterior}.

The approximated posterior densities make sense semantically: in order for the images to differ by 100 pixels,
$\bx_{\epsilon}$ needs to be generated with the coordinates and the radius, which revolve around, but
tend to diverge from $\btheta^*$. To further verify the accuracy of our approach, we draw
10,000 $\btheta$ samples from the resulting Markov chain, simulate respective observations
$\bx_{\epsilon} \sim\ p(\bx|\btheta)$ and each time record the error $\epsilon$ in
relation to the true observation $\bx_o$. Finally, we calculate the average and the standard deviation
from the sample of recorded $\epsilon$ values, which turn out to be 101.46 and 3.52 respectively, confirming
the accuracy of EG-LF-MCMC.

\section{Conclusions and Future Work}
We have presented a novel method for likelihood-free inference, named error-guided likelihood-free MCMC.
It is demonstrated that EG-LF-MCMC outperforms ABC in the ability to infer the true
generating model parameter $\btheta^*$ on problems with semantically and structurally different high-dimensional
observational data. The circle simulator generates deterministic images, while the linear model produces
stochastic numerical samples. Furthermore, we have presented the ability of EG-LF-MCMC to amortize
the inference over variable $\epsilon$ values. To the best of our knowledge, this feature is novel
in the likelihood-free inference literature, and it may enable novel opportunities for various
scientific applications.

In the next phases of our research we plan to analytically study the connection between $p(\btheta|\epsilon)$
and the respective $p(\btheta|\bx)$. We hypothesize that they are equivalent when $\epsilon=0$ and in other cases
$p(\btheta|\epsilon)$ approximates $p(\btheta|\bx)$.

\section*{Acknowledgment}
The computational results presented have been achieved in part using the Vienna Scientific Cluster (VSC).

\bibliographystyle{splncs04}
\bibliography{refs}

\end{document}